\documentclass{article}

\usepackage{microtype}
\usepackage{graphicx}
\usepackage{subfigure}
\usepackage[T1]{fontenc}    
\usepackage{hyperref}       
\usepackage{algorithm}
\usepackage[shortlabels]{enumitem}
\usepackage{wrapfig}
\usepackage{algorithmic} 
\usepackage{natbib}
\usepackage{amsmath}
\usepackage{xcolor}
\usepackage{amsthm}
\usepackage{url}            
\usepackage{booktabs}       
\usepackage{amsfonts}       
\usepackage{nicefrac}       
\usepackage{microtype}      

\usepackage{hyperref}

\newcommand{\ifun}{\iota}

\usepackage[accepted]{icml2018}

\usepackage[textwidth=20mm,textsize=scriptsize]{todonotes}

\icmltitlerunning{MIWAE: Deep Generative Modelling and Imputation of Incomplete Data Sets}

\begin{document}

\twocolumn[
\icmltitle{MIWAE: Deep Generative Modelling and Imputation of Incomplete Data Sets}




\begin{icmlauthorlist}
\icmlauthor{Pierre-Alexandre Mattei}{itu}
\icmlauthor{Jes Frellsen}{itu}
\end{icmlauthorlist}

\icmlaffiliation{itu}{Department of Computer Science, IT University of Copenhagen, Denmark}

\icmlcorrespondingauthor{Pierre-Alexandre Mattei}{pima@itu.dk}
\icmlcorrespondingauthor{Jes Frellsen}{jefr@itu.dk}

\icmlkeywords{Missing data}

\vskip 0.3in
]



\printAffiliationsAndNotice{} 

\begin{abstract}
We consider the problem of handling missing data with deep latent variable models (DLVMs). First, we present a simple technique to train DLVMs when the training set contains missing-at-random data. Our approach, called MIWAE, is based on the importance-weighted autoencoder (IWAE), and maximises a potentially tight lower bound of the log-likelihood of the observed data. Compared to the original IWAE, our algorithm does not induce any additional computational overhead due to the missing data. We also develop Monte Carlo techniques for single and multiple imputation using a DLVM trained on an incomplete data set. We illustrate our approach by training a convolutional DLVM on a static binarisation of MNIST that contains 50\% of missing pixels. Leveraging multiple imputation, a convolutional network trained on these incomplete digits has a test performance similar to one trained on complete data. On various continuous and binary data sets, we also show that MIWAE provides accurate single imputations, and is highly competitive with state-of-the-art methods.
\end{abstract}

\section{Introduction}

In data analysis, we often encounter incomplete data sets. For instance, we may choose to observe a limited set of values due to some constraints; sometimes a data acquisition machine will break; more and more often, respondents do not respond to parts of surveys \citep{national2013nonresponse}. This ubiquity, illustrated for example by \citet{josse2018} or by the monograph of \citet{little2002}, calls for a sound statistical response.

Over the last decades, numerous statistical approaches to handling incomplete data sets have been proposed---reviewed, e.g., by \citet{schafer1997}, \citet{little2002}, and \citet{van2018}. One way of handling the missing values is to replace them by a single set of imputations. Such a procedure, called \emph{single imputation}, can be performed efficiently using nonparametric methods, in particular  $k$-nearest neighbours ($k$NN, see e.g.~\citealp{troyanskaya2001}) or random forests \citep{stekhoven2011}. However, when dealing with high dimensional and complex data distributions \emph{multiple imputation} methods (recently reviewed by \citealp{murray2018}) are often preferred. Such techniques create several sets of imputed values, and consequently allow to quantify the uncertainty of the imputed values. Often, they rely on fitting a parametric model by integrating the missing values out of the likelihood (in a maximum-likelihood setting) or the posterior distribution (in a Bayesian context). Quite often, these approaches are valid under the assumption that the data are \emph{missing-at-random} (MAR, see e.g.~\citealp{little2002}), which essentially means that the missingness mechanism is independent of the missing values.

Earlier attempts mainly relied on using multivariate Gaussian or Student's $t$ distributions (see e.g.~\citealp[Chapters 11 \& 12]{little2002}), with potentially low-rank constraints (e.g.~\citealp{audigier2016}). More complex models have also been explored, including mixtures (e.g.~\citealp{di2007}) and nonparametric Bayes (e.g.~\citealp{si2013}).

\subsection{Deep generative models and incomplete data}

The main purpose of this work is to present a simple way to fit a deep latent variable model (DLVM, \citealp{kingma2014,rezende2014}) to an incomplete data set, and to show how to use it for single or multiple imputation.

DLVMs are latent variable models that draw their high flexibility from deep neural architectures. They can be seen as nonlinear extensions of factor analysis and have links with robust principal component analysis \citep{dai2018} and nonparametric mixture models \citep{mattei2018}. In recent years, DLVMs have had remarkable successes in modelling images~\citep{narayanaswamy2017}, molecules~\citep{kusner2017,gomez2018,riesselman2018}, arithmetic expressions~\citep{kusner2017}, and single-cell  data~\citep{gronbech2018,lopez2018}.

Since DLVMs are generative models, data imputation is a natural application and was actually already considered in the seminal paper of \citet{rezende2014}. However, most of the inferential machinery used within DLVM cannot readily handle incomplete data sets. Indeed, DLVMs are usually fit by maximising lower bounds of the likelihood of the training data. Such bounds include the popular variational autoencoder bound (VAE, \citealp{kingma2014,rezende2014}), and the tighter importance weighted autoencoder (IWAE, \citealp{burda2016}) bound. All these bounds assume that the training data are fully observed. Moreover, the conditionals of a DLVM are generally intractable. Some adaptations are therefore required for DLVMs to handle incomplete data sets efficiently. Specifically, the requirements that we would like to meet are:
\begin{enumerate}[(a)]
    \item Having a sound statistical procedure for training DLVMs on incomplete data sets, under some assumptions on the nature of the missingness mechanism.
    \item Not losing the scalability of variational inference algorithms for DLVMs because of the presence of missing data. In particular, we wish to be able to do online learning on incomplete data sets.
    \item Having sound statistical procedures for both single imputation and multiple imputation.
\end{enumerate}

\subsection{Previous and concurrent work}

Recently, several attempts at training DLVMs in the presence of missing data have been proposed. \citet{yoon2018} considered a way to do so by building on the generative adversarial networks (GANs) of \citet{goodfellow2014}. Their training procedure involves a min-max game which, in the idealised nonparametric limit, allows to recover the data-generating process under the missing-completely-at-random (MCAR, see e.g.~\citealp{little2002}) assumption. The MCAR assumption is stronger than the MAR assumption, and asserts that the missing mechanism is independent of the data. Another GAN-based approach using the MCAR assumption is the one of \citet{li2018}. They optimise jointly three objectives, with respect to six neural networks, which appears quite computationally complex, especially given the optimisation difficulties often noticed when training mix-max objectives like the GANs ones.

An interesting aspect of maximum likelihood inference is that it requires only the weaker MAR assumption. Approximate maximum likelihood using variational inference appears therefore like a natural solution to Requirement (a). \citet{nazabal2018} proposed to slightly change the VAE objective in order to obtain a lower bound of the likelihood of the observed data. Their approach, like ours, uses zero imputation, and their bound is a particular case of ours. They also propose efficient ways of dealing with heterogeneous data sets. An important limitation of their bound is that it cannot be tight, even under the conditions that guarantee tightness of the original VAE bound (these conditions are discussed for example by \citealp{cremer2018}). On the other hand, the likelihood bound that we introduce can be tight, in the limit of very large computational power, similarly to the original IWAE bound that we build on. \citet{ma2018,ma2018eddi}, like \citet{nazabal2018}, revisit the VAE bound in a missing data context. Rather than relying on zero imputation, they elegantly use a more general framework based on permutation invariant neural networks. It is however unclear whether or not their bound can be tight.
Regarding (c), these VAE-based approaches use heuristic schemes that may lead to poor results when the posterior approximation used for variational inference is far from the actual posterior.

A natural way to meet Requirement (c) would be to use the conditional distribution of the missing data. Given a trained DLVM, some Markov chain Monte Carlo (MCMC) schemes have been developed to sample from these conditionals \citep{rezende2014,mattei2018}. However, they require to have access to a complete training set, which is very limiting. A way to design a model with easily approximable conditionals was recently proposed by \citet{ivanov2018variational}. Although their primary concern was to treat cases that involve complete training sets (like image inpainting), they propose a heuristic modification of their objective to handle missing data during training.

\subsection{Contributions}

In this paper, we address Requirement (a) in Section \ref{sec:miwae} by building a simple modification of the IWAE objective of \citet{burda2016}, the \emph{missing data importance-weighted autoencoder (MIWAE) bound}. This objective is a lower-bound of the likelihood of the observed data that can be tight in the limit of very large computational power. Optimising MIWAE, by doing approximate maximum likelihood on the observed data, is sound in a MAR scenario. The computational cost of our algorithm does not exceed the one of the original IWAE, therefore satisfying Requirement (b).

Moreover, once a DLVM is trained using MIWAE, we show how to use it for missing data imputation. Specifically, we show how to approximate the conditional expectation of any function of the missing data using importance sampling---a particular application is single imputation. We also show how to perform multiple imputation using sampling importance resampling. While approximate, these two algorithms, presented in Section \ref{sec:MC}, will give the right answer in the limit of an infinite number of samples, therefore addressing Requirement (c) in a grounded way.

\section{Training DLVMs with missing data}
\label{sec:miwae}
We start with some i.i.d.~data stored in a matrix $\mathbf{X} = (\mathbf{x}_1,\ldots,\mathbf{x}_n)^T \in \mathcal{X}^n$. We assume that $p$ different features are present in the data, leading to the decomposition $\mathcal{X} = \mathcal{X}_1\times \ldots \times \mathcal{X}_p$ (for example, if all features are continuous, $\mathcal{X} =\mathbb{R}^p$).

When some data are missing, we split each sample $i \in \{1,\ldots,n\}$ into the observed features $\mathbf{x}_i^\text{o}$ and the missing features $\mathbf{x}_i^\text{m}$. The indices of the missing features are stored in binary vectors $\mathbf{m}_i \in \{ 0,1\}^p$ such that $m_{ij}=0$ if feature $j$ is observed for sample $i$, and $m_{ij}=1$ if feature $j$ is missing.

We wish to explain these potentially high-dimensional data using some latent variables $\mathbf{Z} = (\mathbf{z}_1,\ldots,\mathbf{z}_n)^T \in \mathbb{R}^{n \times d}$. While the dimension $d$ of the latent space will often be smaller than the dimension of the input space $\mathcal{X}$, this will not always be the case.

\subsection{Deep latent variable models}

DLVMs \citep{kingma2014,rezende2014} assume that $(\mathbf{x}_i,\mathbf{z}_i)_{i\leq n}$ are driven by the following generative model:
\begin{equation}
\left\{
\begin{array}{ll}
\mathbf{z} \sim p(\mathbf{z}) \\
p_{\boldsymbol{\theta}}(\mathbf{x}|\mathbf{z}) = \Phi(\mathbf{x}| f_{\boldsymbol{\theta}}(\mathbf{z})),
\end{array}
\right.
\end{equation}
where
\begin{itemize}
	\item $p(\mathbf{z})$ is the \emph{prior distribution of the latent variables}, which is often a standard Gaussian distribution (e.g.~\citealt{kingma2014,rezende2014}), but might be much more complex and learnable (e.g.~\citealt{tomczak2018}).
	\item $(\Phi (\cdot|\boldsymbol{\eta}))_{\boldsymbol{\eta} \in H}$ is a parametric family of distributions over $\mathcal{X}$ called the \emph{observation model}. Often, it is a very simple family such as the Gaussian distribution if $\mathcal{X}$ is continuous, products of multinomial distributions if $\mathcal{X}$ is binary. It can even simply be the family of Dirac distributions---this is for example the case for generative adversarial networks \citep{goodfellow2014} or flow-based generative models (e.g.~\citealt{dinh2017,kingma2018}).
	\item $f_{\boldsymbol{\theta}}: \mathbb{R}^d \rightarrow H$ is a function called the \emph{decoder} (or the \emph{generative network}), and is parametrised by a deep neural network whose weights are stored in $\boldsymbol{\theta} \in \boldsymbol{\Theta}$. Prior knowledge about the problem at hand may be distilled into the architecture of $f_{\boldsymbol{\theta}}$: for example, by using convolutional layers when dealing with images (e.g.~\citealt{salimans2017}) or recurrent networks when dealing with sequential data (e.g.~\citealt{bowman2016,gomez2018}).
\end{itemize}

\textbf{Assumptions.} We make the assumption that the data are MAR, which is crucial to legitimate inference strategies based on the observed likelihood. Moreover, we make the fairly general assumption that the observation model $(\Phi (\cdot|\boldsymbol{\eta}))_{\boldsymbol{\eta} \in H}$ is such that all its marginal distributions are available in closed-form, which is true for commonly used observation models like the Gaussian or the Bernoulli ones. This assumption will conveniently guarantee that the quantity $p_{\boldsymbol{\theta}}(\mathbf{x}^\text{o}|\mathbf{z})$ is easy to compute. We do not make any further assumption about the form of the prior or the architecture of the decoder.

\subsection{Amortised Variational Inference with missing data}

DLVMs are usually trained using approximate maximum likelihood techniques that maximise lower bounds of the log-likelihood function. Under the MAR assumption, the relevant quantity to maximise is the likelihood of the observed data $\mathbf{x}_1^\text{o},\ldots,\mathbf{x}_n^\text{o} $, which is equal to
\begin{equation}
\label{eq:likelihood}
\small
\ell (\boldsymbol{\theta}) =  \sum_{i=1}^n \log  p_{\boldsymbol{\theta}}(\mathbf{x}_i^\text{o}) =  \sum_{i=1}^n  \log \int p_{\boldsymbol{\theta}}(\mathbf{x}_i^\text{o}|\mathbf{z})p(\mathbf{z})d\mathbf{z}.
\end{equation}
Unfortunately, the integrals involved in this expression make direct maximum likelihood impractical. However, it is possible to derive tractable tight lower bounds of $\ell (\boldsymbol{\theta})$ whose maximisation is much easier. More precisely, we will base our inference strategy on the IWAE approach of \citet{burda2016}, which combines ideas from \emph{amortised variational inference} \citep{gershman2014,kingma2014,rezende2014} and \emph{importance sampling} (see e.g.~\citealt{gelman2013}, Section 10.4).

To build a lower bound using amortised variational inference, we will define a parametrised conditional distribution $q_{\boldsymbol{\gamma}}(\mathbf{z}|\mathbf{x}^\text{o})$ called the \emph{variational distribution} that will \emph{play the role of a proposal distribution close to the intractable posterior} $p_{\boldsymbol{\theta}}(\mathbf{z}|\mathbf{x}^\text{o})$.

Specifically, this conditional distribution will be defined as
\begin{equation}
q_{\boldsymbol{\gamma}}(\mathbf{z}|\mathbf{x}^\text{o}) = \Psi(\mathbf{z} | g_{\boldsymbol{\gamma}} (\ifun(\mathbf{x}^\text{o})),
\end{equation} where
\begin{itemize}
	\item  $\ifun$ is an \emph{imputation function chosen beforehand} that transforms $\mathbf{x}^\text{o}$ into a complete input vector $\ifun(\mathbf{x}^\text{o}) \in \mathcal{X}$ such that $\ifun(\mathbf{x}^\text{o})^\text{o} = \mathbf{x}^\text{o}$.
	\item The set $(\Psi(\cdot | \boldsymbol{\kappa}))_{\boldsymbol{\kappa} \in \mathcal{K}}$ is a parametric family of simple distributions over $\mathbb{R}^d$, called the \emph{variational family}.
	\item The function $g_{\boldsymbol{\gamma}}: \mathcal{X} \rightarrow \mathcal{K}$, called the \emph{inference network} or the \emph{encoder}, is parametrised by a deep neural network whose weights are stored in $\boldsymbol{\gamma} \in \mathbf{\Gamma}$. Its role will be to transform each data point into the parameters of $\Psi$.
\end{itemize}

Following the steps of  \citet{burda2016}, we can use the distribution $q_{\boldsymbol{\gamma}}$ to build approachable stochastic lower bounds of $\ell(\boldsymbol \theta)$. More specifically, given $K \in \mathbb{N}^*$, we define the  \emph{missing data importance-weighted autoencoder (MIWAE) bound}

\begin{multline}
\label{eq:missIWAE}
\mathcal{L}_K (\boldsymbol{\theta,\gamma}) = \\ \sum_{i=1}^n \mathbb{E}_{\mathbf{z}_{i1},\ldots,\mathbf{z}_{iK} \sim q_{\boldsymbol{\gamma}}(\mathbf{z}|\mathbf{x}^\text{o}_i)} \left[ \log\frac{1}{K} \sum_{k=1}^K \frac{p_{\boldsymbol{\theta}}(\mathbf{x}_i^\text{o}|\mathbf{z}_{ik})p(\mathbf{z}_{ik})}{q_{\boldsymbol{\gamma}}(\mathbf{z}_{ik}|\mathbf{x}^\text{o}_i)} \right].
\end{multline}
Jensen's \citeyearpar{jensen1906} inequality ensures that $\mathcal{L}_K (\boldsymbol{\theta,\gamma})\leq \ell(\theta)$, which means that $\mathcal{L}_K$ is indeed a lower bound of the observed log-likelihood.
This inequality and the analogy with importance sampling motivate the name MIWAE for $\mathcal{L}_K$.

When $K=1$, the bound resembles the variational autoencoder (VAE, \citealp{kingma2014,rezende2014}) objective, so we call this bound the \emph{MVAE bound}. In very interesting concurrent work, \citet{nazabal2018} derived independently the MVAE bound. They also discuss efficient strategies to design observation models that handle heterogeneous data sets.

The MIWAE bound is exactly the expected value of what we would get if we were to estimate the log-likelihood by approximating the integrals $ p_{\boldsymbol \theta}(\mathbf{x}_1),\ldots, p_{\boldsymbol \theta}(\mathbf{x}_n)$ present in Eqn.~\eqref{eq:likelihood} using importance sampling, with proposal distributions $q_{\boldsymbol{\gamma}}(\mathbf{z}|\mathbf{x}_1^\text{o}),\ldots,q_{\boldsymbol{\gamma}}(\mathbf{z}|\mathbf{x}_n^\text{o})$. Regarding these importance sampling problems, it follows from a classical result of Monte Carlo integration (see e.g.~\citealt[Section 3.3.2]{robert2004}) that the optimal (in the sense of minimising the variance of the estimate) proposal distributions would exactly be the posterior distributions $p_{\boldsymbol{\theta}}(\mathbf{z}|\mathbf{x}_1^\text{o}),\ldots,p_{\boldsymbol{\theta}}(\mathbf{z}|\mathbf{x}_n^\text{o})$. For this reason, for all $i \in \{1,\ldots,n\}$, we may interpret (and use) $q_{\boldsymbol{\gamma}}(\mathbf{z}|\mathbf{x}_i^\text{o})$  as an approximation of the posterior $p_{\boldsymbol{\theta}}(\mathbf{z}|\mathbf{x}_i^\text{o})$. The nature of this approximation is investigated further in section~\ref{sec:properties}.

\subsection{Building Blocks of the MIWAE bound}

The use of the imputation function $\ifun$, which fills in the missing values in each data point in order to feed it to the encoder, is the main originality of the MIWAE bound. Intuitively, it would be desirable to use an accurate imputation function. However, thanks to the properties of importance sampling, we will see in the next subsection that \emph{using a very rough imputation is acceptable}. In particular, in all the experiments of this paper, the function $\ifun$ \emph{will simply be the zero imputation function}, which replaces missing values by zeroes. \citet{nazabal2018} also proposed to use zero imputation together with the MVAE bound. More complex imputations can of course be used. In particular it would also be possible to use a parametrised imputation function that could be learned by maximising $\mathcal{L}_K$. An interesting idea in that direction was recently proposed in a VAE context by \citet{ma2018eddi}, who advocate the use of a permutation invariant neural network. Interestingly, the zero imputation function is a particular case of their proposal. However, in this paper, we argue that the use of importance sampling considerably alleviate the drawbacks of having a rough imputation function, such as the zero imputation one.

Regarding the variational family, the original IWAE paper of \citet{burda2016} and used diagonal Gaussians (following \citealp{kingma2014} and \citealp{rezende2014}). However, it is worth noticing that, similarly to the IWAE, our MIWAE bound could be used for any reparametrisable variational family, like the ones listed by \citet{figurnov2018}. Moreover, an interesting alternative to Gaussians would be to consider elliptical distributions, as recently proposed and advocated by \citet{domke2018}. Indeed, such distributions allow to model different tail behaviours for the proposals $q_{\boldsymbol{\gamma}}(\mathbf{z}|\mathbf{x}_i^\text{o})$, which is desirable in an importance sampling context. Specifically, in our experiments on UCI data sets, we used products of Student's t distributions. 

In general, the encoder can simply be a fully connected network, as in \citet{burda2016}. However, when dealing with images or sounds, it will often be more reasonable to use a convolutional network, as in \citet{salimans2017}. When the data are sequential, recurrent networks are also an option (see e.g.~\citealt{bowman2016,gomez2018}).

\subsection{Properties of the MIWAE bound}
\label{sec:properties}

The main difference between our bound and the original IWAE bound of \citet{burda2016} is that ours lower bounds the likelihood of an incomplete data set---using a proposal distribution that approximates $p_{\boldsymbol{\theta}}(\mathbf{z}|\mathbf{x}^\text{o})$---while the original IWAE objective is a lower bound of the complete likelihood---whose proposal approximates $p_{\boldsymbol{\theta}}(\mathbf{z}|\mathbf{x})$. When there is no missing data, our bound reduces to the original IWAE bound. \emph{The MIWAE bound can thus be seen as a simple generalisation of the IWAE bound that can accommodate to missing data.} As a result, the properties of the MIWAE bound described in this subsection are directly obtained by leveraging recent theoretical work on the IWAE bound \citep{cremer2017,nowozin2018,rainforth18,domke2018,tucker2018}.

The one-sample MVAE bound $\mathcal{L}_1$ may be rewritten
\begin{equation}
\label{eq:KL1}
\mathcal{L}_1(\boldsymbol{\theta},\boldsymbol{\gamma}) = \ell (\boldsymbol{\theta}) - \text{KL}\left(\prod_{i=1}^n q_{\boldsymbol{\gamma}}(\mathbf{z}_i|\mathbf{x}_i^\text{o}) \Big|\Big| \prod_{i=1}^n p_{\boldsymbol{\theta}}(\mathbf{z}_i|\mathbf{x}_i^\text{o})\right).
\end{equation}
Therefore, maximising $\mathcal{L}_1$ will lead to minimising the Kullback-Leibler divergence between the true posterior distribution of the codes, and the variational distribution. In that sense, the variational approximation obtained by maximising $\mathcal{L}_1$ can be seen as a Kullback-Leibler projection of the posterior distribution on a set of distributions restricted by the choice of the variational family and the architecture of the encoder. For perspectives on the balance between variational family and encoder architecture, see \citet{cremer2018}.

When $K>1$, however, the variational approximation that maximises $\mathcal{L}_K$ cannot be seen as a Kullback-Leibler projection of the posterior. Indeed, as studied by \citet{cremer2017}, \citet{naesseth2018}, and \citet{domke2018} the variational distribution $q_{\boldsymbol{\gamma}}$ in \eqref{eq:KL1} is replaced by a more complex distribution $q_{\text{IW}}$ which depends both on $\boldsymbol{\theta}$ and $\boldsymbol{\gamma}$. This fact has both a good consequence---the bound will be tighter since $q_{\text{IW}}$ is more complex---and a bad one---$q_{\boldsymbol{\gamma}}(\mathbf{z}|\mathbf{x}^\text{o})$ might become a poor approximation of $p_{\boldsymbol{\theta}}(\mathbf{z}|\mathbf{x}^\text{o})$ as $K$ grows. \citet{rainforth18} give a formal treatment of this ambivalence, and propose modifications of the IWAE objective that tackle it.  \citet[Theorem 3]{domke2018} noted that, when $K \rightarrow \infty$, the variational approximation that maximises $\mathcal{L}_K$ may rather be seen as a $\chi$ projection rather than a Kullback-Leibler one. These properties are justifications for our use of $q_{\boldsymbol{\gamma}}(\mathbf{z}|\mathbf{x}_i^\text{o})$  as an approximation of the posterior $p_{\boldsymbol{\theta}}(\mathbf{z}|\mathbf{x}_i^\text{o})$ in the Monte Carlo schemes that we introduce in Section \ref{sec:MC}.

Importantly, \citet{burda2016} proved that, the larger $K$, the tighter the bound. A direct consequence of their Theorem 1 is that when the posterior distributions have lighter tails than their variational counterparts,
\begin{equation}
\mathcal{L}_1(\boldsymbol{\theta},\boldsymbol{\gamma}) \leq \mathcal{L}_2(\boldsymbol{\theta},\boldsymbol{\gamma}) \leq \ldots \leq \mathcal{L}_K(\boldsymbol{\theta},\boldsymbol{\gamma}) \xrightarrow[K \rightarrow \infty]{ } \ell (\boldsymbol{\theta}).
\end{equation}
Under some mild moments conditions on the importance weights, the Theorem 3 of \citet{domke2018} will ensure that, when $K\rightarrow \infty$,
\begin{equation}
\mathcal{L}_K(\boldsymbol{\theta},\boldsymbol{\gamma}) = \ell(\boldsymbol{\theta})+O(1/K),
\end{equation}
where the constant in $O(1/K)$ is linked to the $\chi$ divergence between $q_{\boldsymbol{\gamma}}(\mathbf{z}|\mathbf{x}_i^\text{o})$ and the true posterior.
These results show that, provided that we use a large $K$, optimising the MIWAE bound will lead to optimising a tight bound of the likelihood, even if the imputation function $\ifun$ gives poor results. This motivates our choice to use simple imputation schemes for $\ifun$ that can scale easily to large data sets. Specifically, in all the experiments of this paper, we chose to use the very naive but scalable zero imputation function. We will confirm empirically that, as long as $K$ exceeds a few dozens, the accuracy of the imputation function has little effect on training.

In practice, the expectations in Eqn.~\eqref{eq:missIWAE} are hard to compute, but we can obtain unbiased estimates of the gradients of the bound $\mathcal{L}_K$ by sampling from the variational distribution. While this was the original approach of \citet{burda2016}, \citet{roeder2017} and \citet{rainforth18} noticed both experimentally and theoretically that such gradients may have an overly large variance. A solution to this problem, based on a double use of the reparametrisation trick, was recently proposed by \citet{tucker2018}. While we did not implement this trick, it is very likely that using it would improve the optimisation of our MIWAE bound.

\section{Missing data imputation}

\label{sec:MC}

We have derived a statistically sound objective for training DLVMs in the presence of missing data. However, it is often useful to be also able to impute these data, through either single or multiple imputation. From now on, we assume that we have already trained our DLVM using the MIWAE bound, resulting in two known distributions $p_{\boldsymbol{\theta}}$ and $q_{\boldsymbol{\gamma}}$. We assume in this section that we are given a data point $\mathbf{x} \in \mathcal{X}$ composed of some observed features  $\mathbf{x}^{\text{o}}$ and some missing data $\mathbf{x}^{\text{m}}$. This data point can be taken from the training set, or from another incomplete data set sampled from the same distribution.

\subsection{Existing Techniques for Imputation with DLVMs}

Since we have already learned a generative model $p_{\boldsymbol{\theta}}$, a good way of imputing $\mathbf{x}^{\text{m}}$ would be to sample according to its conditional distribution \begin{equation}
p_{\boldsymbol{\theta}}(\mathbf{x}^{\text{m}}|\mathbf{x}^{\text{o}}) = \int p_{\boldsymbol{\theta}}(\mathbf{x}^{\text{m}}|\mathbf{x}^{\text{o}},\mathbf{z})p(\mathbf{z}|\mathbf{x}^{\text{o}})d\mathbf{z}.
\end{equation}
The nonlinearity of the decoder makes this conditional distribution hard to assess. However, it is possible to sample from a Markov chain whose stationary distribution is exactly $p_{\boldsymbol{\theta}}(\mathbf{x}^{\text{m}}|\mathbf{x}^{\text{o}})$. \citet{rezende2014} first proposed a simple scheme to generate samples that approximatively follow $p_{\boldsymbol{\theta}}(\mathbf{x}^{\text{m}}|\mathbf{x}^{\text{o}})$ by starting with some random imputation, which will be autoencoded several times. Their approach was then improved by \citet{mattei2018}, who derived a Metropolis-within-Gibbs sampler that asymptotically produces samples from  $p_{\boldsymbol{\theta}}(\mathbf{x}^{\text{m}}|\mathbf{x}^{\text{o}})$.

These two techniques were developped to use DLVMs trained on complete data, and critically require the availability of a good approximation of the posterior distribution of the complete data $p_{\boldsymbol{\theta}}(\mathbf{z}|\mathbf{x})$. Indeed, if $p_{\boldsymbol{\theta}}(\mathbf{z}|\mathbf{x})$ is far from $q_{\boldsymbol{\gamma}}(\mathbf{z}|\mathbf{x})$, the scheme of \citet{rezende2014} will converge to a distribution far away from $p_{\boldsymbol{\theta}}(\mathbf{x}^{\text{m}}|\mathbf{x}^{\text{o}})$, while the Metropolis-within-Gibbs sampler of \citet{mattei2018} will converge to the right distribution but at an impractical convergence rate. 

However, while training a VAE or an IWAE will provide an approximation of $p_{\boldsymbol{\theta}}(\mathbf{z}|\mathbf{x})$ via the inference network, using the MIWAE bound rather provides an approximation of $p_{\boldsymbol{\theta}}(\mathbf{z}|\mathbf{x}^{\text{o}})$---see Section\ref{sec:properties}. Therefore, the schemes of both \citet{rezende2014} and \citet{mattei2018} seem unfit for our purposes. We will therefore derive a new imputation technique compatible with a DLVM trained using the MIWAE bound.

\subsection{Single Imputation with MIWAE}

The main idea of our imputation scheme is to leverage the fact that $q_{\boldsymbol{\gamma}}(\mathbf{z}|\mathbf{x}^{\text{o}})$ is a good approximation of $p_{\boldsymbol{\theta}}(\mathbf{z}|\mathbf{x}^{\text{o}})$.

Let us first focus on the \emph{single imputation problem}: finding a single imputation $\hat{\mathbf{x}}^{\text{m}}$ that is close to the true $\mathbf{x}^{\text{m}}$. If the data are continuous and the $\ell_2$ norm is a relevant error metric, then the optimal decision-theoretic choice would be
$
\hat{\mathbf{x}}^{\text{m}} = \mathbb E [\mathbf{x}^{\text{m}} | \mathbf{x}^{\text{o}}]
$, which is likely to be intractable for the same reasons $p_{\boldsymbol{\theta}}(\mathbf{x}^{\text{m}}|\mathbf{x}^{\text{o}})$ is. We can actually give a recipe to estimate the more general quantity
\begin{multline}
\mathbb E [h(\mathbf{x}^{\text{m}}) | \mathbf{x}^{\text{o}}]
=\int h(\mathbf{x}^{\text{m}})p_{\boldsymbol{\theta}}(\mathbf{x}^{\text{m}}|\mathbf{x}^{\text{o}})d\mathbf{x}^{\text{m}} \\
= \iint h(\mathbf{x}^{\text{m}})p_{\boldsymbol{\theta}}(\mathbf{x}^{\text{m}}|\mathbf{x}^{\text{o}},\mathbf{z})p_{\boldsymbol{\theta}}(\mathbf{z}|\mathbf{x}^{\text{o}})d\mathbf{z}d\mathbf{x}^{\text{m}},
\end{multline}
where $h(\mathbf{x}^{\text{m}})$ is any absolutely integrable function of $\mathbf{x}^{\text{m}}$. Indeed, this integral can be estimated using self-normalised importance sampling with the proposal distribution $p_{\boldsymbol{\theta}}(\mathbf{x}^{\text{m}}|\mathbf{x}^{\text{o}},\mathbf{z})q_{\boldsymbol{\gamma}}(\mathbf{z}|\mathbf{x}^{\text{o}})$, leading to the estimate
\begin{equation}
\label{eq:IS}
\mathbb E [h(\mathbf{x}^{\text{m}}) | \mathbf{x}^{\text{o}}] \approx \sum_{l=1}^L w_l h\left(\mathbf{x}^{\text{m}}_{(l)}\right),
\end{equation}
where $(\mathbf{x}^{\text{m}}_{(1)},\mathbf{z}_{(1)}),\ldots,(\mathbf{x}^{\text{m}}_{(L)},\mathbf{z}_{(L)})$ are i.i.d.~samples from $p_{\boldsymbol{\theta}}(\mathbf{x}^{\text{m}}|\mathbf{x}^{\text{o}},\mathbf{z})q_{\boldsymbol{\gamma}}(\mathbf{z}|\mathbf{x}^{\text{o}})$ that can be sampled via simple ancestral sampling, and, for all $l \in \{1,\ldots,L\}$
\begin{equation}
\label{eq:weights}
w_l=\frac{r_l}{r_1+...+r_L}, \; \text{with} \; r_l = \frac{p_{\boldsymbol{\theta}}(\mathbf{x}^{\text{o}}|\mathbf{z}_{(l)})p(\mathbf{z}_{(l)})}{q_{\boldsymbol{\gamma}}(\mathbf{z}_{(l)}|\mathbf{x}^{\text{o}})}.
\end{equation}
This estimate can be computed by performing a single encoding of $\ifun(\mathbf{x}^{\text{m}})$ and $L$ decodings of the samples from the variational distribution. This is an additional advantage compared to the schemes of \citet{rezende2014} and \citet{mattei2018}, which require one encoding and one decoding at each iteration of the Markov chain.

Often the quantity $\mathbb{E}[h(\mathbf{x}^{\text{m}}) | \mathbf{x}^{\text{o}},\mathbf{z}]$ is available in closed-form. This is for example the case when $h$ is the identity function and the observation model has closed-form conditional means (like multivariate Gaussians or products of univariate distributions). In that situation, a slightly improved estimate is
\begin{equation}
\mathbb E [h(\mathbf{x}^{\text{m}}) | \mathbf{x}^{\text{o}}] \approx \sum_{l=1}^L w_l \mathbb{E}\left[h(\mathbf{x}^{\text{m}}) | \mathbf{x}^{\text{o}},\mathbf{z}_{(l)}\right],
\end{equation}
where the weights are defined as in Eqn.~\eqref{eq:weights}. By virtue of the law of total variance, the variance of this new estimate will always be smaller than the one defined in Eqn.~\eqref{eq:IS}.

\subsection{Multiple Imputation with MIWAE}

Often, it is more interesting to have access to a large number of i.i.d.~samples from the conditional distribution than to estimate its moments. Such \emph{multiple imputation} is also approachable using the importance sampling scheme of the previous subsection.

Indeed, using \emph{sampling importance resampling} (see e.g.~\citealt[Section 10.4]{gelman2013}) with the weights defined in Eqn.~\eqref{eq:weights} allows to draw some samples that will be approximatively i.i.d.~samples from $p_{\boldsymbol{\theta}}(\mathbf{x}^{\text{m}}|\mathbf{x}^{\text{o}})$ when $L$ is large.

The simple process to sample $M$ samples from the condtional distribution goes at follows. First, for some $L\gg M$, we sample $(\mathbf{x}^{\text{m}}_{(1)},\mathbf{z}_{(1)}),\ldots,(\mathbf{x}^{\text{m}}_{(L)},\mathbf{z}_{(L)})$ and compute the weights similarly to the previous subsection. Then, we sample $M$ imputations with replacement from the set $\{\mathbf{x}^{\text{m}}_{(1)},\ldots,\mathbf{x}^{\text{m}}_{(L)} \}$, the probability of each sample being exactly its importance weight.

\section{Experiments}

In this section, we explore empirically the properties of DLVMs trained with the MIWAE bound, as well as their imputations. Regarding general implementation details, we used TensorFlow probability \citep{dillon2017} and the Adam optimiser \citep{kingma2014adam} to train all DLVMs. All neural nets are initialised following the heuristics of \citet{glorot2010} and \citet{he2015}. Because we wanted to investigate the properties of the different bounds, all DLVMs are trained without any form of regularisation. Regarding gradient estimation, we used the pathway derivative estimate of \citet{roeder2017} for MNIST, but found that it let to quite unstable training for the continuous UCI data sets, so we used the regular gradient estimates of \citet{burda2016}. These results are consistent with the recent findings of \citet{tucker2018}, who showed that the pathway derivative estimate is actually biased, but that this bias is almost invisible for MNIST.

\subsection{Training a convolutional MIWAE on MNIST}

We wish to train a DLVM on an incomplete version of the static binarisation of  MNIST (with $50\%$ of the pixels missing uniformly at random), using the convolutional architecture of \citet{salimans2015markov}. To assess the validity of our claim that using a naive imputation function is not too harmful when $K$ is moderate, we compare using the zero imputation function, and using an oracle imputation that utilises the true values of the missing pixels. To that end, we focus on two models: one trained with the MIWAE bound with $K=50$, and one trained using the MVAE bound (i.e.~$K=1$).
As baselines, we also train a regular VAE and a regular IWAE using the complete data sets. To compare models, we evaluate estimates of their test log-likelihood obtained using importance sampling with $5 000$ samples and an inference network refitted on the test set, as suggested by \citet{cremer2018} and \citet{mattei2018c}.

The results are shown in Figure \ref{fig:likelihood}. In the MVAE case ($K=1$, which is essetially equivalent to the framework of \citealp{nazabal2018}), using the oracle imputation provides a clear improvement. In the MIWAE case (with $K=50$), both imputations schemes are on par (in accordance with our hypothesis), and outperform MVAE. As shown in Figure \ref{fig:likelihood}, MIWAE (with zero imputation) obtained is almost competitive with a VAE trained on the complete MNIST data set. 

\begin{figure}
	\centering
	\includegraphics[width=\columnwidth]{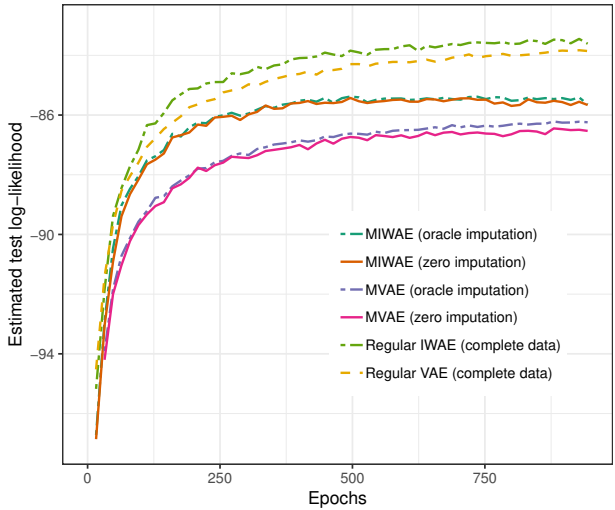}
	\caption{Estimated test log-likelihood of various models trained on binary MNIST as a function of the number of training epochs. The MIWAE model was trained using $K=50$ importance weights.}
	\label{fig:likelihood}
\end{figure}

\textbf{Single imputation.} We also evaluate the quality of the imputations provided by MIWAE. To this end, we use MVAE and MIWAE with zero imputation, together with the proposed importance sampling scheme for imputation, and compare them to a state-of-the-art single imputation algorithm: missForest (\citealp{stekhoven2011}, fit using the \texttt{missForest} R package with default options). The results are displayed in Figure~\ref{fig:imps}. When $L \geq 10$, both MVAE and MIWAE outperform missForest, and MIWAE provides the most accurate imputations when $L \geq 1\,000$. Some incomplete and imputed digits from the trainig set are also presented in Figure~\ref{fig:missimp}.

\begin{figure}
	\centering	
	\includegraphics[width=\columnwidth]{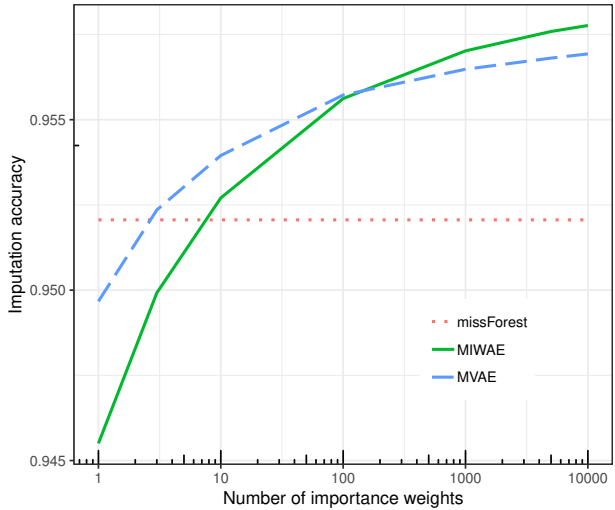}	
	\caption{Test imputation accuracy as a function of the number of importance weights ($L$) used in the single imputation scheme for various models trained on binary MNIST. The MIWAE model was trained using $K=50$ importance weights.}
	\label{fig:imps}	
\end{figure}

\textbf{Multiple imputation.}
To evaluate multiple imputation, we consider the task of classifying the incomplete binary MNIST data set.
We train a two-layer convolutional network (whose architecture is similar to the dropout one of \citealp{Wan2013} and model selection is done using early stopping) using the original data and and some imputed versions, and assess the classification performance on the test set. The results are shown in Table~\ref{tab:multiple}. Regarding MIWAE, we use both the single imputation obtained with $10\,000$ importance samples, and a multiple imputation of 20 complete data sets obtained using sampling importance resampling.
Interestingly, \emph{the convolutional network trained with the 20 imputed data sets outperforms the one trained on complete data} in terms of classification error (but not when it comes to the test cross-entropy). This suggests that the DLVM trained using MIWAE generalises quite well, and may also be used efficiently for data augmentation.

\begin{table}[b]
	\centering
	\small
	\begin{tabular}{llllll} \toprule
		&  \emph{Test accuracy}       & \emph{Test cross-entropy} \\ \midrule
		Zero imp.                   & 0.9739 (0.0018) & 0.1003 (0.0092) \\
		missForest imp.    & 0.9805 (0.0018) & 0.0645 (0.0066) \\
		MIWAE single imp.         & 0.9847 (0.0009) & 0.0510 (0.0035) \\
		MIWAE multiple imp.     & \textbf{0.9868 (0.0008)} & 0.0509 (0.0044) \\
		Complete data                 & 0.9866 (0.0007) & \textbf{0.0464 (0.0026)} \\ \bottomrule
	\end{tabular}
	\caption{Test accuracy and cross-entropy obtained by training a convolutional network using the imputed versions of the static binarisation of MNIST. The numbers are the mean of 10 repeated trainings with different seeds and standard deviations are shown in brackets.}
	\label{tab:multiple}
\end{table}

\begin{figure*}[t!]
	\begin{center}
		\includegraphics[width=2.\columnwidth]{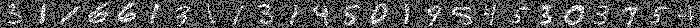}
		\includegraphics[width=2.\columnwidth]{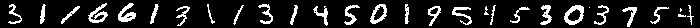}'
		\includegraphics[width=2.\columnwidth]{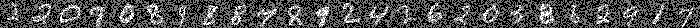}
		\includegraphics[width=2.\columnwidth]{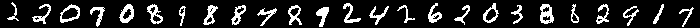}
	\end{center}
	\caption{Random incomplete samples from the MNIST training data set, and the imputations obtained by MIWAE (trained with $K=50$ importance weights, and imputed with $L=10\,000$ importance weights).}
	\label{fig:missimp}
\end{figure*}

\begin{table*}[t]
	\begin{center}
	\begin{tabular}{llllllll} \toprule
		& \emph{Banknote} & \emph{Breast}        & \emph{Concrete} & \emph{Red} & \emph{White} & \emph{Yeast} \\ \midrule
		MIWAE    & \textbf{0.446 (0.038)} & \textbf{0.280 (0.021)}  & \textbf{0.501 (0.040)} &\textbf{0.643 (0.026)} &\textbf{0.735 (0.033)}& \textbf{0.964(0.057)} \\
		MVAE         & 0.593 (0.059) & 0.318 (0.018) &0.587(0.026)& 0.686 (0.120) &0.782 (0.018)& 0.997 (0.064)\\
		missForest         & 0.676 (0.040) & 0.291 (0.026)  &0.510 (0.11)& 0.697 (0.050) &0.798 (0.019)& 1.41 (0.02)\\
		PCA     & 0.682 (0.016) & 0.729 (0.068)  &0.938 (0.033)& 0.890 (0.033)&0.865 (0.024) & 1.05(0.061)\\
		$k$NN                 & 0.744 (0.033) & 0.831 (0.029) & 0.962(0.034)& 0.981 (0.037) &0.929 (0.025)&1.17 (0.048)\\
		Mean                 &1.02 (0.032) &1.00 (0.04)  & 1.01 (0.035)& 1.00 (0.03)&1.00 (0.02)&1.06 (0.052)\\ \bottomrule
	\end{tabular}
		\end{center}
	\caption{Mean-squared error for single imputation for various continuous UCI data sets (mean and standard deviations over $5$ randomly generated incomplete data sets).}
	\label{tab:uci}
\end{table*}

\subsection{Single imputation of UCI data sets}

We now address the single imputation of various continuous data sets from the UCI database \citep{dua2017}. All data sets are corrupted by removing half of the features uniformly at random.
As competitors, we chose the mean imputation, which is an often inaccurate baseline, the $k$NN approach of \citet{troyanskaya2001} fit using the \texttt{VIM} R package \citep{kowarik2016}, with the number of neighbours chosen in $\{5,...,15\}$
using $5$-fold cross validation; the missForest approach of \citet{stekhoven2011} (fit using the \texttt{missForest} R package with default options); a technique based on principal component analysis (PCA, \citealp{josse2012handling}), fit using the \texttt{missMDA} R package \citep{josse2016}, where the number of principal components is found using generalised cross validation \citep{josse2012selecting}.

We also trained
two DLVMs with the same general features: the intrinsic dimension $d$ is fixed to 10, which may be larger than the actual number of features in the data, but DLVMs are known to automatically ignore some latent dimensions \citep{dai2018}; both encoder and decoder are
multi-layer perceptrons with $3$ hidden layers (with $128$ hidden units) and tanh activations; we use products of Student's $t$ for the variational family (following \citealp{domke2018}) and the observation model (following \citealp{ijcai}). We perform $500\,000$ gradient steps for all data sets; no regularisation scheme is used, but the observation model is constrained so that the eigenvalues of its covariances are larger than $0.01$ (as suggested by \citealp{mattei2018}), when the data are scaled, this means that the code explains at most 99\% of the variance of the features. 

One of the DLVM was trained using the MVAE bound, which is equivalent to the objective of \citet{nazabal2018}, and one using the MIWAE bound with $K=20$. 

We used both models to impute the data sets by estimating the conditional means with $L=10\,000$ importance samples.


The results, averaged over $5$ randomly corrupted repetitions, are presented in Table \ref{tab:uci}. MIWAE provides more accurate imputations than all competitors. Interestingly, going from $K=1$ (i.e., MVAE) to $K=20$ significantly improves the accuracy of the imputations. None of the other recent DLVM-based approaches is within the competitors of this experiment. However, note that \citet{ivanov2018variational} reimplemented the GAN-based approach of \citet{yoon2018} to compare it with their VAE-based model. They concluded that none of them significantly outperformed missForest, which may still be considered a state-of-the-art algorithm. We used the same architecture for all data sets; it is therefore likely that architecture search would significantly improve these results.

\section{Conclusion and future work}

We have proposed a simple way of performing approximate maximum likelihood training for a DLVM with an incomplete data set, by maximising an objective called the MIWAE bound. We have also shown how to use the obtained model for single and multiple imputation.

A first way of improving the MIWAE optimisation would be to use the low variance gradients recently proposed by \citet{tucker2018} for similar objectives. Another avenue would be to study more refined imputation functions, even though we have shown that zero imputation does a very good job together with MIWAE. For instance, we could try to use the permutation invariant networks advocated by \citet{ma2018,ma2018eddi}.
An important limitation is that MIWAE can only deal with MAR data. Trying to extend to the missing-not-at-random case might be possible by explicitly modelling the missing data mechanism.

\bibliography{biblio}
\bibliographystyle{icml2018}

\end{document}
